\documentclass[conference,compsoc]{IEEEtran}
\ifCLASSOPTIONcompsoc
  \usepackage[nocompress]{cite}
\else
  \usepackage{cite}
\fi
\ifCLASSINFOpdf
  \usepackage[pdftex]{graphicx}
\else
\fi
\usepackage{amsmath}
\usepackage{hyperref}
\usepackage{xcolor}
\usepackage{csvsimple}
\usepackage{longtable}
\usepackage{array}
\usepackage{booktabs}

\begin{document}
%

\title{Automatic generation of reviews of scientific papers}



%
\author{\IEEEauthorblockN{Anna Nikiforovskaya\IEEEauthorrefmark{1}\IEEEauthorrefmark{2},
Nikolai Kapralov\IEEEauthorrefmark{3},
Anna Vlasova\IEEEauthorrefmark{4}, 
Oleg Shpynov\IEEEauthorrefmark{5}\IEEEauthorrefmark{6} and
Aleksei Shpilman\IEEEauthorrefmark{1}\IEEEauthorrefmark{5}}
\IEEEauthorblockA{\IEEEauthorrefmark{1}National Research University Higher School of Economics, Saint Petersburg, Russia}
\IEEEauthorblockA{\IEEEauthorrefmark{2}IDMC, Université de Lorraine, Nancy, France}
\IEEEauthorblockA{\IEEEauthorrefmark{3}Sechenov Institute of Evolutionary Physiology and Biochemistry RAS, Saint Petersburg, Russia}
\IEEEauthorblockA{\IEEEauthorrefmark{4}Saint Petersburg State University, Saint Petersburg, Russia}
\IEEEauthorblockA{\IEEEauthorrefmark{5}JetBrains Research, Saint Petersburg, Russia}
\IEEEauthorblockA{\IEEEauthorrefmark{6}Corresponding author, Email: os@jetbrains.com}
}


\maketitle

\begin{abstract}
With an ever-increasing number of scientific papers published each year, it becomes more difficult for researchers to explore a field that they are not closely familiar with already. This greatly inhibits the potential for cross-disciplinary research. A traditional introduction into an area may come in the form of a review paper. However, not all areas and sub-areas have a current review. In this paper, we present a method for the automatic generation of a review paper corresponding to a user-defined query. This method consists of two main parts. The first part identifies key papers in the area by their bibliometric parameters, such as a graph of co-citations. The second stage uses a BERT based architecture that we train on existing reviews for extractive summarization of these key papers. We describe the general pipeline of our method and some implementation details and present both automatic and expert evaluations on the PubMed dataset.

\end{abstract}


%
\IEEEpeerreviewmaketitle

\section{Introduction}

When approaching a subject that they are not familiar with, a researcher often starts with a search of relevant papers. For example, Google Scholar and Scopus are commonly used tools to search for scientific papers~\cite{tool_comparison}. Besides the search by title, author names, or keywords, these search engines also provide the user with different statistics, like citations of the paper. However, more often than not, it is hard to organize the information from these papers, especially with the current rise in publication numbers. For instance, regarding the latest COVID-19 epidemic, almost 50000 papers were produced in the last six months.

Some automatic tools approach the task from the bibliometric point of view. One example is bibliometric maps that can be built~\cite{bibl_comparison}. These maps visualize the way scientific papers are related in the chosen scientific area using extra information about the paper, like its authors, place of publication, and papers it is cited in. Bibliometric methods allow for highlighting the most important or interesting papers in the selected area. Bibliometric methods often use citations as a measure of scientific impact or paper importance in the area. However, it was shown that the author and the place of publication of the paper affect the number of citations~\cite{dong2016can}. Also, the meaning of citation is actively studied. For example, it has been shown that there are 15 different meanings of the citation~\cite{aksnes2019citations}. 

Another aspect of the paper analysis is the summarization of the scientific papers~\cite{automa_summ}. Studies in this area show that the citation context, i.e., the text surrounding the link to the paper, can be used for its summarization~\cite{mohammad2009using}. Moreover, it was demonstrated that citation context reflects the meaning of the paper better than the paper's abstract, and that citation themselves can be used for paper summarization~\cite{qazvinian2010citation}. However, to our knowledge, no studies exist that present an approach that would summarize a group of papers on a specific topic.

When the area of research is established, a researcher might instead refer to a textbook or a review paper. These sources have information already processed and neatly organized. The enormous amount of papers in popular research areas makes it hard to explore the scientific achievements and to discover possible future research directions. Therefore, automatic tools for scientific papers analysis are demanded by researchers and hence are being developed in various areas.

In this paper, we present a novel method for automatic review generation that combines bibliometric analysis of a specific research area to identify key papers together with a BERT-based deep neural network trained to extract the most relevant sentence from these key papers. The result is a tool able to automatically generate a review based on a query from the user. 

We also asked experts in various biological areas to evaluate our tool applied to the PubMed database\cite{pubmed-corpus}, and the results demonstrate that generated reviews are indeed relevant to posted queries. 

The rest of the paper is organized as follows. First, we describe relevant work in the areas of bibliometry and automatic summarization. Then we detail our method, along with the preprocessing process we applied to the PubMed database. Lastly, we present the results of automatic and expert evaluation and conclude the paper.

\section{Related Work}

\subsection{Bibliometric analysis of scientific papers}

Historically, bibliometrics has arisen from the statistical studies of bibliographies\cite{egghe1990introduction}. Nowadays, it can be applied to all sorts of publications, from scientific papers and books\cite{de2009bibliometrics} to newspapers and patents\cite{madani2016evolution}. Statistics, such as the identification of authors with the highest number of publications and countries with the highest contribution to the research field, can be obtained from such analysis. 

Most of the bibliometric methods are based on paper similarity, which can be determined by co‐citation analysis, bibliographic coupling, direct citation, and a bibliographic coupling‐based citation‐text hybrid approach. Moreover, a citation graph can be used to investigate the total citations number of a paper and its dynamics. Algorithms like PageRank\cite{page1999pagerank} are capable of detecting the most exciting or revolutionary articles, that can change the direction of studies in any particular field. Co-citation graphs are used for cluster analysis and identifying dense communities of similar papers. Bibliographic coupling uses a number of common citations as similarity metrics between papers. Hybrid methods imply all of these and have been shown in recent studies to be the most prominent ones since they allow us to detect similar papers and represent the research front in an unbiased manner \cite{boyack2010co}. 

Different tools are available for bibliometric analysis: standalone desktop applications like VosViewer\cite{van2011text} or packages for various programming languages such as Bibliometrics package \cite{aria2017bibliometrix} for the R programming language. Another group is websites like Google Scholar offering search services on a particular subject and the ability to perform citation analysis. Dedicated citations indexes databases, like Scopus or Web of Science, can be used to export bibliographic data for a batch of papers. Altogether, existing bibliometric tools either require manual data processing or provide limited analysis capabilities. 

The majority of tools outputs paper's abstract to a user. Reading several dozens of abstracts that often describe the results of the paper in broad terms and contain additional, often redundant information may be cumbersome for a user. We believe that sentences from reviews that cite the paper might be a better alternative. However, since not all papers have associated reviews, we need a way for the automatic generation of review-like sentences.

\subsection{Extractive summarization}
Text summarization approaches can be abstractive or extractive. Abstractive summarization attempts to produce summaries close to the ones a human expert would make. Those summaries might contain new phrases and sentences, which do not occur in the original text. Extractive summarization produces the text composited from sentences of the original text. In this study, we use extractive summarization, as it guarantees that the generation process would not corrupt the information. In this way, we can find the exact sentence in the paper if we need to see its context

Most extractive summarization methods can be presented as a three-step process~\cite{automa_summ}. First, they create an intermediate representation of the sentences, then they score sentences based on that representation, and finally generate the summary based on those scores. 


Lately, deep learning methods are used more and more, including in the task of extractive summarization \cite{cheng2016neural, nallapati2017summarunner}. These methods use recurrent neural networks or convolutional neural networks to evaluate sentences. In the process of training, these networks create vector representations of sentences. It is also important to note that vector representations can be acquired by training the network to solve an auxiliary task, such as creating a language model~\cite{li2018word}. This helps in the case of limited data availability.

In 2018, a new deep neural network-based language model named BERT was introduced~\cite{devlin2018bert}. Using this model for natural language processing tasks has led to significant improvements in the results. In particular, in 2019, it was shown that BERT modification for extractive summarization (BERTSUM) is superior in quality to standard machine learning methods and previously available deep learning methods \cite{liu2019fine}. We have chosen this method as our base model and describe it and our modifications in section~\ref{sec:ext}. Figure~\ref{bertsum-cmp} shows the architecture of this method.

\begin{figure}[ht]
    \centering
    \includegraphics[width=\linewidth]{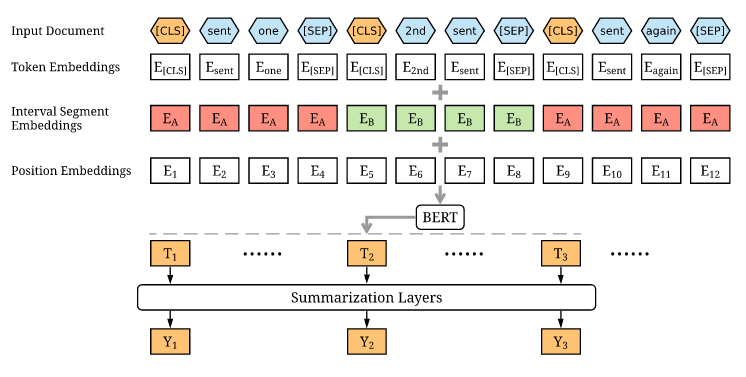}
    \caption{BERTSUM deep summarization network (from \cite{liu2019fine}).}
    \label{bertsum}
\end{figure}

Early experiments in the summarization of scientific papers mainly concentrate on abstract generation~\cite{paice1993identification} by using the structure of the article and the general rules for dividing papers into sections.

However, it was later shown that \textit{citation contexts} of the paper better reflect the paper content, especially if the information from them is selected properly~\cite{abu2011coherent}. \textit{Citation contexts} are sentences from other papers that describe the contents of the target paper. These sentences can be easily identified by the presence of the link to the target paper.


Most works in summarization evaluate the generated summary by comparing it to the reference summary, written by the expert (i.e., the paper abstract). Two commonly used metrics to compare the generated summary and the reference summary are ROUGE and BLEU. They are both based on n-grams count. However, it was shown that in summarization tasks, ROUGE better correlates with human evaluation than BLEU~\cite{lin2003automatic}. ROUGE metric is defined as follows:

$$\text{ROUGE-n} = \frac{\sum\limits_{S\in \{\text{reference texts}\}} \sum\limits_{gram_i \in S} Count_{match}(gram_i)}{\sum\limits_{S\in \{\text{reference texts}\}} \sum\limits_{gram_i \in S} Count(gram_i)}\text{, }$$
	
where $Count_{match}(gram_i)$ is the number of times n-gram $gram_i$ occurs both in the reference text $S$ and in the generated text, while $Count(gram_i)$ is the number of times n-gram $gram_i$ occurs in reference text $S$. In our study we use $\text{ROUGE} = \frac12(\text{ROUGE-1} + \text{ROUGE-2})$ to compare generated summaries and reference ones. 

Automatic evaluation techniques rarely show human satisfaction with the generated summaries. To this end, human evaluation is often employed for a better quality control~\cite{automa_summ, xiong2014empirical}. In our work, we use both automatic and human expert evaluation of the generated summary.

\section{Methods}

\begin{figure*}[ht!]
    \centering
    \includegraphics[width=\textwidth]{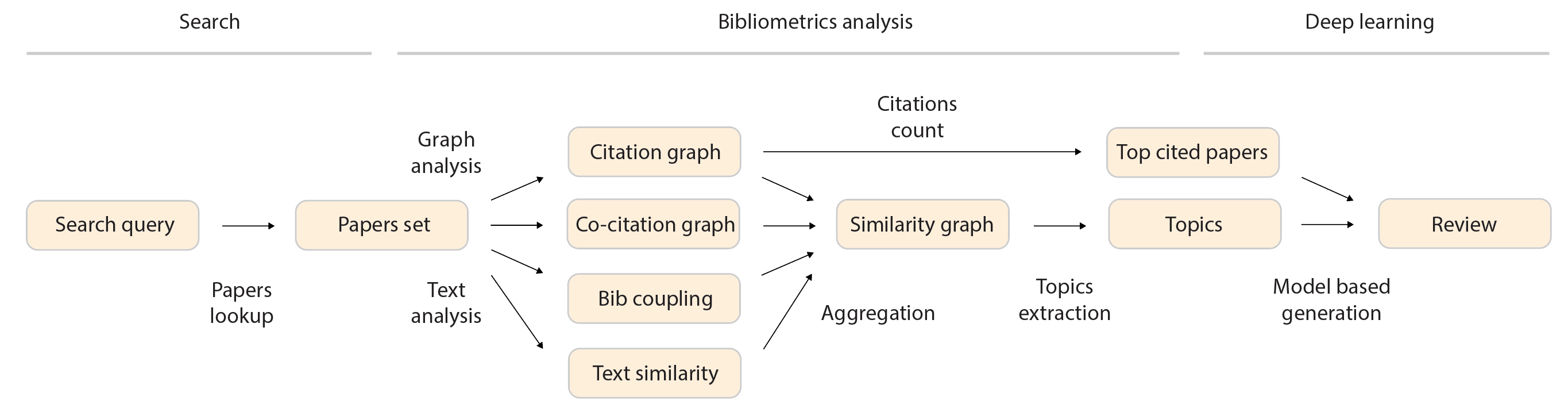}
    \caption{Overall pipeline of our automatic review generation method.}
    \label{pipeline}
\end{figure*}

Figure \ref{pipeline} shows the general pipeline of our automatic review generation method. In this section, we will describe every stage of that process in detail, but we start from the description of the data used.

\subsection{Data preprocessing}

Our method can be used together with any sufficiently large corpora of scientific papers collected for research in bibliometry and related areas~\cite{automa_summ, hunter2006biomedical}. 

For this work, we chose PubMedCentral Author Manuscript Collection~\cite{pubmed-corpus} that contains over 600~000 papers, mostly in areas of biology and medicine. Articles are stored in Journal Article Tag Suite XML format~\cite{jats-format}. The data entry includes the paper's unique identifier (PMID) title, abstract, main text, table and figures captions, authors, PMIDs of cited papers, and publication. All this additional information can help in this study, which makes this corpus the most convenient.

Each paper in the corpus has a type. It can be either a general research paper or a review paper, which summarizes some specific scientific areas. There are 8000 review papers out of 600~000 total papers.




We preprocess the raw data into the following data tables. Each has a PMID (a unique identifier of the paper) as key:

\begin {itemize}
    \item Lists of sentences of each paper that we can use to quickly find all sentences of a paper or sentences under certain numbers. It contains sentence's number in the paper, and the text of the sentence itself. 
    \item Abstracts.
    \item Image caption texts. 
    \item Table caption texts. 
    \item List of review paper PMIDs. Lists PMIDs of all review papers that cite this paper.
    \item Citations. PMIDs of cited papers. 
\end{itemize}

An additional table with ``reverse citations", that allows for a quick search of citation context by cited paper. With these tables, we can easily access a paper and all associated papers by merging related tables by a specific PMID.




\subsection{Citation graph analysis}

We developed a tool that combines the capabilities of paper search engines with bibliometrics analysis. The service crawls the PubMed and Semantic Scholar databases as bibliographic data providers and keeps up-to-date information in the Neo4j graph database with daily updates. 

The automatic review pipeline starts with a search query by keywords or phrases, where users can choose the most recent, the most cited, or the most relevant papers and limit the size of search results. A citation graph is built on-the-fly and shows overall publication dynamics in time and is used to detect the most popular articles, authors, and journals. We use a hybrid approach for papers similarity determination - a combination of citation, co-citation graph features, bibliographic coupling, and text-based similarity based on the TF-IDF metric\cite{ramos2003using}. This approach yields excellent results in the case of well-established research areas where citations and co-citations graphs contain a lot of information, as well as in arising topics such as Covid-19, where citations graph is almost empty due to the recency of the papers. 

We employ one of the most popular algorithms for community detection - Louvain\cite{raghavan2007near} algorithm to extract closely related groups of papers (topics) from the similarity graph. For each topic, the service shows a word cloud of topic-specific keywords and detailed information about included papers.  


Analysis results are combined and presented as an automatically generated review compiled from most review-like sentences from target papers. This review provides a bird-eye view of the research area. It is presented as a table with sentences, quality scores, and additional information about original articles, including topic, publication year, citation number, and a digital online identifier.

\subsection{Review-based extractive summarization}\label{sec:ext}

As previously mentioned, we base our extractive summarization method on BERTSUM introduced by Liu Yang~\cite{liu2019fine}. The model introduced takes in several sentences separated by specific tags ``[CLS]'' and ``[SEP]'' that mark the beginning and end of the fragments of interest, i.e., the sentences. For each input sentence $i$ the BERTSUM model produces a vector $T_i$, which will be further perceived as a vector representation of the original sentence. Then these vectors are transferred to the linear layer, which will give an estimate of the quality of the sentence for the summary $Y_i$.

The original model is then trained with binary cross-entropy, as the whole problem was perceived as a binary classification problem, and the goal was to get binary values $Y_i$. In our work, we state the problem differently, as we do not have an ideal set of sentences. We perceive a problem as a regression problem by trying to predict the review-like quality of the sentence, and therefore we have changed the way of training the model.

This way, we consider review papers as ``ideal" summaries. However, review papers do not contain the sentences of the cited papers. We can then compare the sentences from the paper with its citation context in review papers to decide if it should be in the summary. We calculate the ROUGE score $R_i$ between the sentence $i$ of the paper and its citation context in review papers. We then use this score as a target value for training the model.

The new loss function during model training is then defined as follows:

\begin{equation}
    \text{MSE}(Y, R) = \frac1n \sum\limits_{i = 1}^{n} (Y_i - R_i)^2
\end{equation}

A scientific paper's text is generally much longer than BERT can process during one iteration. BERT model can accept 10-15 sentences per iteration. Due to this fact, we have changed the way of training the model. The text is split into intersecting blocks with the length of the intersection set to 5 sentences. 





\section{Results}

In this section, we present the automatic evaluation of our modified BERTSUM summarization method by examining ROUGE scores of extracted sentences compared to sentences in review papers. We also show the example of a pipeline's output and perform human evaluation by asking experts in specific research areas to grade the output of our method. 

\subsection{Automatic evaluation}

\begin{figure}[ht]
    \centering
    \includegraphics[width=\linewidth]{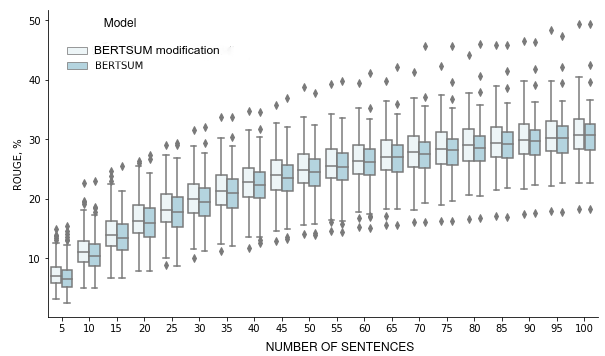}
    \caption{Comparison of non-modified and modified BERTSUM. Our modifications are described in section~\ref{sec:ext}.}
    \label{bertsum-cmp}
\end{figure}

We base the automatic evaluation on the way we state the summarization problem. Since we postulated review papers as ``perfect'' summaries, we created a test datasets from 300 review papers for benchmarking the summarization. The final summary is selected by choosing $n$ sentences, given the model's highest scores. Those result summaries are compared to the original review paper via the ROUGE score. The higher the number of sentences, the easier it is to summarize the paper. Hence the ROUGE score generally increases with the increase of $n$.

We have compared the original BERTSUM and our modified version and present the results in figure~\ref{bertsum-cmp}. Our modified method shows improvement over base BERTSUM in terms of ROUGE score, especially in cases of low $n$. Since our goal is to create a concise review down to one sentence, if possible, results demonstrate that our modifications help improve the pipeline's overall performance.

\subsection{Example output of the pipeline}

Table~\ref{tab:ex-result} shows an example of output sentences associated with the query ``Alzheimer's disease". For every paper, we present the best sentence in terms of the model score is. This score represents how close the sentence to the possible citation context of the paper, i.e., the sentence from a hypothetical review.

As can be seen from the table, our method outputs a diverse collection of papers, covering different aspects of the area of interest. Summary sentences also reflect essential points presented in their corresponding papers, thus helping a potential user to evaluate the state of the research at a glance. 

\begin{table*}[!htb]
    \centering
    \renewcommand{\arraystretch}{2}
\begin{tabular}{|c|c|c|c|}%
    \hline
    \bfseries Paper & \bfseries Year & \bfseries Summary sentence & \bfseries Score\\
    \cmidrule{1-4} \parbox{6cm}{Epidemiology of Alzheimer disease.} &2012 & \parbox{7cm}{The global prevalence of dementia has been estimated to be as high as 24 million, and is predicted to double every 20 years until at least 2040.} & 0.108\\
\cmidrule{1-4} \parbox{6cm}{Neuropathological alterations in Alzheimer disease.} &2011 & \parbox{7cm}{Importantly, these cross-sectional neuropathological data have been largely validated by longitudinal in vivo studies using modern imaging biomarkers such as amyloid PET and volumetric MRI.} & 0.093\\
\cmidrule{1-4} \parbox{6cm}{Cognitive reserve and Alzheimer disease.} &2006 & \parbox{7cm}{There is also the possibility that directly enhancing CR may help forestall the diagnosis of AD.} & 0.087\\
\cmidrule{1-4} \parbox{6cm}{Oxidative stress and Alzheimer disease.} &2000 & \parbox{7cm}{Research in the field of molecular biology has helped to provide a better understanding of both the cascade of biochemical events that occurs with Alzheimer disease (AD) and the heterogeneous nature of the disease.} & 0.091\\
\cmidrule{1-4} \parbox{6cm}{Systemic inflammation and disease progression in Alzheimer disease.} &2009 & \parbox{7cm}{OBJECTIVE To determine whether acute episodes of systemic inflammation associated with increased TNF-alpha would be associated with long-term cognitive decline in a prospective cohort study of subjects with Alzheimer disease.} & 0.115\\
\cmidrule{1-4} \parbox{6cm}{Classification and basic pathology of Alzheimer disease} &2009 & \parbox{7cm}{Extracellular A$\beta$ accumulation occurs in the parenchyma as diffuse, focal or stellate deposits.} & 0.085\\
\cmidrule{1-4} \parbox{6cm}{Loneliness and risk of Alzheimer disease.} &2007 & \parbox{7cm}{CONCLUSION Loneliness is associated with an increased risk of late-life dementia but not with its leading causes.} & 0.071\\
\cmidrule{1-4} \parbox{6cm}{The Genetics of Alzheimer Disease: Back to the Future} &2010 & \parbox{7cm}{Three decades of genetic research in Alzheimer disease (AD) have substantially broadened our understanding of the pathogenetic mechanisms leading to neurodegeneration and dementia.} & 0.099\\
\cmidrule{1-4} \parbox{6cm}{Increased risk of type 2 diabetes in Alzheimer disease.} &2004 & \parbox{7cm}{Both type 2 diabetes (35\% vs. 18\%; P \&lt; 0.05) and IFG (46\% vs. 24\%; P \&lt; 0.01) were more prevalent in Alzheimer disease versus non-Alzheimer disease control subjects, so 81\% of cases of Alzheimer disease had either type 2 diabetes or IFG.} & 0.107\\
\cmidrule{1-4} \parbox{6cm}{Hyperinsulinemia and risk of Alzheimer disease.} &2004 & \parbox{7cm}{The HR of AD for the highest quartile of insulin compared to the lowest was 1.7 (95\% CI: 1.0, 2.7; p for trend = 0.009).} & 0.073\\

\bottomrule
\end{tabular}
    \vspace{1mm}
	\caption{Example output for query ``Alzheimer's disease".}
    \label{tab:ex-result}
\end{table*}

\subsection{Expert evaluation}

To assess the quality of the result summaries in terms of relevance to the topic of the query and usefulness to the researcher, we have conducted a human expert evaluation on five queries given by experts in specific fields. For each query, the most important 20 papers were selected based on their citations. Then, from each article, one sentence was selected using our modified BERTSUM extractive summarization model.

We asked the experts to classify the generated summary sentences into the three following groups:

\begin{itemize}
    \item \textbf{Not relevant} -- the sentence is not relevant to the query.
    \vspace{3mm}
    \item \textbf{Relevant} -- the sentence is relevant to the query, but does not provide useful information about the area.    
    \vspace{3mm}
    \item \textbf{Useful} --  the suggestion is useful for understanding the scientific area.
\end{itemize}

\begin{table}[!htb]
    \centering
    \renewcommand{\arraystretch}{2}
    \begin{tabular}{|c|c|c|c|}
	\hline
	    \textbf{\small{Query}} & \textbf{\small{Not Relevant}} & \textbf{\small{Relevant}} & \textbf{\small{Useful}}  \\
	    \cmidrule{1-4}
	    \parbox{3cm}{klebsiella pneumoniae plasmid virulence} & 5\% & 55\%  & 40\%\\
	    \cmidrule{1-4}
	    \parbox{3cm}{neisseria meningitidis serogroup} & 10\% & 35\% & 55\% \\
	    \cmidrule{1-4}
	    \parbox{3cm}{psi prion propagation in yeast} & 10\% & 20\% & 70\% \\
	    \cmidrule{1-4}
	    \parbox{3cm}{brain computer interface} & 5\% & 25\% & 70\% \\
	    \cmidrule{1-4}
	    \parbox{3cm}{brain stimulation} & 5\% & 30\% & 65\% \\
	    \cmidrule{1-4}
	    Mean Values: & 7\% & 33\% & 60\%\\
	    \bottomrule
	\end{tabular}
	\vspace{1mm}
	\caption{Results of the human expert evaluation.}
    \label{tab:expert-test}
\end{table}

We present the result of this evaluation in table~\ref{tab:expert-test}. As you can see, the fraction of irrelevant sentences generated is manageably small an 7\%, while the average rate of relevant and useful sentences are 33\% and 60\% respectively. This demonstrates that our method of automatic review generation with extractive summarization produces a diverse review with sentences that can give insight into the queried area of research.

\section{Conclusion}
In this work, we present a method of automatic review generation by combining bibliometric analysis for key papers identification and deep learning natural language processing by a BERT-based network that evaluates how review-worthy are sentences in identified key papers. The resulting tool generates a list of sentences, each one best describing the result of key papers in response to a query from the user. We evaluate our tool automatically on the PubMed dataset by ROUGE score and manually by asking experts in the field to evaluate specific queries. Both evaluations show that our method can produce relevant one-sentence descriptions of papers.

This tool could be employed to significantly increase researchers' ability to process information in a novel area. It may also assist in writing a traditional review paper or a textbook chapter on a subject. 

The code for our method is available at \url{https://github.com/JetBrains-Research/pubtrends-review}.






\bibliographystyle{IEEEtran}
\bibliography{paper.bib}

\begin{thebibliography}{10}
\providecommand{\url}[1]{#1}
\csname url@samestyle\endcsname
\providecommand{\newblock}{\relax}
\providecommand{\bibinfo}[2]{#2}
\providecommand{\BIBentrySTDinterwordspacing}{\spaceskip=0pt\relax}
\providecommand{\BIBentryALTinterwordstretchfactor}{4}
\providecommand{\BIBentryALTinterwordspacing}{\spaceskip=\fontdimen2\font plus
\BIBentryALTinterwordstretchfactor\fontdimen3\font minus
  \fontdimen4\font\relax}
\providecommand{\BIBforeignlanguage}[2]{{%
\expandafter\ifx\csname l@#1\endcsname\relax
\typeout{** WARNING: IEEEtran.bst: No hyphenation pattern has been}%
\typeout{** loaded for the language `#1'. Using the pattern for}%
\typeout{** the default language instead.}%
\else
\language=\csname l@#1\endcsname
\fi
#2}}
\providecommand{\BIBdecl}{\relax}
\BIBdecl

\bibitem{tool_comparison}
M.~E. Falagas, E.~I. Pitsouni, G.~A. Malietzis, and G.~Pappas, ``Comparison of
  pubmed, scopus, web of science, and google scholar: strengths and
  weaknesses,'' \emph{The FASEB Journal}, vol.~22, no.~2, pp. 338--342, 2008.

\bibitem{bibl_comparison}
M.~Cobo, A.~López-Herrera, E.~Herrera-Viedma, and F.~Herrera, ``Science
  mapping software tools: Review, analysis, and cooperative study among
  tools,'' \emph{Journal of the American Society for Information Science and
  Technology}, 2011.

\bibitem{dong2016can}
Y.~Dong, R.~A. Johnson, and N.~V. Chawla, ``Can scientific impact be
  predicted?'' \emph{IEEE Transactions on Big Data}, vol.~2, no.~1, pp. 18--30,
  2016.

\bibitem{aksnes2019citations}
D.~W. Aksnes, L.~Langfeldt, and P.~Wouters, ``Citations, citation indicators,
  and research quality: An overview of basic concepts and theories,''
  \emph{Sage Open}, vol.~9, no.~1, p. 2158244019829575, 2019.

\bibitem{automa_summ}
A.~Nenkova and K.~McKeown, ``Automatic summarization,'' \emph{Foundations and
  Trends in Information Retrieval}, vol.~5, pp. 103--233, 2011.

\bibitem{mohammad2009using}
S.~Mohammad, B.~Dorr, and M.~E. et~al., ``Using citations to generate surveys
  of scientific paradigms,'' in \emph{Proceedings of human language
  technologies: The 2009 annual conference of the North American chapter of the
  association for computational linguistics}, 2009, pp. 584--592.

\bibitem{qazvinian2010citation}
V.~Qazvinian, D.~Radev, and A.~{\"O}zg{\"u}r, ``Citation summarization through
  keyphrase extraction,'' in \emph{Proceedings of the 23rd international
  conference on computational linguistics (COLING 2010)}, 2010, pp. 895--903.

\bibitem{pubmed-corpus}
\BIBentryALTinterwordspacing
 [Online]. Available:
  \url{https://www.ncbi.nlm.nih.gov/pmc/about/mscollection/}
\BIBentrySTDinterwordspacing

\bibitem{egghe1990introduction}
L.~Egghe and R.~Rousseau, \emph{Introduction to informetrics: Quantitative
  methods in library, documentation and information science}.\hskip 1em plus
  0.5em minus 0.4em\relax Elsevier Science Publishers, 1990.

\bibitem{de2009bibliometrics}
N.~De~Bellis, \emph{Bibliometrics and citation analysis: from the science
  citation index to cybermetrics}.\hskip 1em plus 0.5em minus 0.4em\relax
  scarecrow press, 2009.

\bibitem{madani2016evolution}
F.~Madani and C.~Weber, ``The evolution of patent mining: Applying
  bibliometrics analysis and keyword network analysis,'' \emph{World Patent
  Information}, vol.~46, pp. 32--48, 2016.

\bibitem{page1999pagerank}
L.~Page, S.~Brin, R.~Motwani, and T.~Winograd, ``The pagerank citation ranking:
  Bringing order to the web.'' Stanford InfoLab, Tech. Rep., 1999.

\bibitem{boyack2010co}
K.~W. Boyack and R.~Klavans, ``Co-citation analysis, bibliographic coupling,
  and direct citation: Which citation approach represents the research front
  most accurately?'' \emph{Journal of the American Society for information
  Science and Technology}, vol.~61, no.~12, pp. 2389--2404, 2010.

\bibitem{van2011text}
N.~J. Van~Eck and L.~Waltman, ``Text mining and visualization using
  vosviewer,'' \emph{arXiv preprint arXiv:1109.2058}, 2011.

\bibitem{aria2017bibliometrix}
M.~Aria and C.~Cuccurullo, ``bibliometrix: An r-tool for comprehensive science
  mapping analysis,'' \emph{Journal of informetrics}, vol.~11, no.~4, pp.
  959--975, 2017.

\bibitem{cheng2016neural}
J.~Cheng and M.~Lapata, ``Neural summarization by extracting sentences and
  words,'' \emph{arXiv preprint arXiv:1603.07252}, 2016.

\bibitem{nallapati2017summarunner}
R.~Nallapati, F.~Zhai, and B.~Zhou, ``Summarunner: A recurrent neural network
  based sequence model for extractive summarization of documents,'' in
  \emph{Thirty-First AAAI Conference on Artificial Intelligence}, 2017.

\bibitem{li2018word}
Y.~Li and T.~Yang, ``Word embedding for understanding natural language: a
  survey,'' in \emph{Guide to Big Data Applications}.\hskip 1em plus 0.5em
  minus 0.4em\relax Springer, 2018, pp. 83--104.

\bibitem{devlin2018bert}
J.~Devlin, M.-W. Chang, K.~Lee, and K.~Toutanova, ``Bert: Pre-training of deep
  bidirectional transformers for language understanding,'' \emph{arXiv preprint
  arXiv:1810.04805}, 2018.

\bibitem{liu2019fine}
Y.~Liu, ``Fine-tune bert for extractive summarization,'' \emph{arXiv preprint
  arXiv:1903.10318}, 2019.

\bibitem{paice1993identification}
C.~D. Paice and P.~A. Jones, ``The identification of important concepts in
  highly structured technical papers,'' in \emph{Proceedings of the 16th annual
  international ACM SIGIR conference on Research and development in information
  retrieval}, 1993, pp. 69--78.

\bibitem{abu2011coherent}
A.~Abu-Jbara and D.~Radev, ``Coherent citation-based summarization of
  scientific papers,'' in \emph{Proceedings of the 49th Annual Meeting of the
  Association for Computational Linguistics: Human Language Technologies-Volume
  1}.\hskip 1em plus 0.5em minus 0.4em\relax Association for Computational
  Linguistics, 2011, pp. 500--509.

\bibitem{lin2003automatic}
C.-Y. Lin and E.~Hovy, ``Automatic evaluation of summaries using n-gram
  co-occurrence statistics,'' in \emph{Proceedings of the 2003 Human Language
  Technology Conference of the North American Chapter of the Association for
  Computational Linguistics}, 2003, pp. 150--157.

\bibitem{xiong2014empirical}
W.~Xiong and D.~Litman, ``Empirical analysis of exploiting review helpfulness
  for extractive summarization of online reviews,'' in \emph{Proceedings of
  coling 2014, the 25th international conference on computational linguistics:
  Technical papers}, 2014, pp. 1985--1995.

\bibitem{hunter2006biomedical}
L.~Hunter and K.~B. Cohen, ``Biomedical language processing: what's beyond
  pubmed?'' \emph{Molecular cell}, vol.~21, no.~5, pp. 589--594, 2006.

\bibitem{jats-format}
\BIBentryALTinterwordspacing
 [Online]. Available:
  \url{https://groups.niso.org/apps/group_public/download.php/21030/ANSI-NISO-Z39.96-2019.pdf}
\BIBentrySTDinterwordspacing

\bibitem{ramos2003using}
J.~Ramos \emph{et~al.}, ``Using tf-idf to determine word relevance in document
  queries,'' in \emph{Proceedings of the first instructional conference on
  machine learning}, vol. 242.\hskip 1em plus 0.5em minus 0.4em\relax New
  Jersey, USA, 2003, pp. 133--142.

\bibitem{raghavan2007near}
U.~N. Raghavan, R.~Albert, and S.~Kumara, ``Near linear time algorithm to
  detect community structures in large-scale networks,'' \emph{Physical review
  E}, vol.~76, no.~3, p. 036106, 2007.

\end{thebibliography}
%



\end{document}